\documentclass[conference]{IEEEtran}
\IEEEoverridecommandlockouts
\usepackage{cite}
\usepackage{amsmath,amssymb,amsfonts}
\usepackage{algorithmic}
\usepackage{graphicx}
\usepackage{textcomp}
\usepackage{xcolor}
\usepackage{hyperref}
\usepackage{mathtools}
\usepackage{booktabs}
\usepackage{arydshln}
\usepackage{subcaption}
\def\BibTeX{{\rm B\kern-.05em{\sc i\kern-.025em b}\kern-.08em
    T\kern-.1667em\lower.7ex\hbox{E}\kern-.125emX}}
    
\newcommand*{\captionsource}[2]{%
  \caption[{#1}]{%
    #1%
    \textbf{ Source:} #2%
  }%
}

\begin{document}

\title{Contrastive Unpaired Translation using Focal Loss for Patch Classification
}

\author{\IEEEauthorblockN{Bernard Spiegl}
\IEEEauthorblockA{\textit{Faculty of Electrical Engineering and Computing, University of Zagreb}\\
Zagreb, Croatia \\
bernard.spiegl@fer.hr}
}

\maketitle

\begin{abstract}
Image-to-image translation models transfer images from input domain to output domain in an endeavor to retain the original content of the image.
Contrastive Unpaired Translation is one of the existing methods for solving such problems.
Significant advantage of this method, compared to competitors, is the ability to train and perform well in cases where both input and output domains are only a single image.
Another key thing that differentiates this method from its predecessors is the usage of image patches rather than the whole images.
It also turns out that sampling negatives (patches required to calculate the loss) from the same image achieves better results than a scenario where the negatives are sampled from other images in the dataset.
This type of approach encourages mapping of corresponding patches to the same location in relation to other patches (negatives) while at the same time improves the output image quality and significantly decreases memory usage as well as the time required to train the model compared to CycleGAN method used as a baseline.
Through a series of experiments we show that using focal loss in place of cross-entropy loss within the PatchNCE loss can improve on the model's performance and even surpass the current state-of-the-art model for image-to-image translation.
\end{abstract}


\section{Introduction}

In the recent years, deep convolutional networks and deep learning methods have become increasingly popular for tackling a wide variety of problems.
As the demand for different types of data used for training deep convolutional models is continuously on the rise, significant efforts are being put into methods that would allow us to generate completely new, synthetic data from pre-existing data. One particularly interesting method is image-to-image translation which aims to take images from one domain and translate them so they have the characteristics of another domain whilst retaining the contents of the original images.

This paper covers \textit{Contrastive Unpaired Image-to-Image Translation} \cite{park2020cut}, a successor to the \textit{CycleGAN} \cite{CycleGAN2017} method from 2017 and proposes tweaks to the model that result in notable performance increases.
This new method harnesses the use of generative adversarial networks \cite{NIPS2014_5ca3e9b1} while both improving on the training speed and accuracy on certain problems in comparison to other baseline methods (e.g.\ \textit{CycleGAN} \cite{CycleGAN2017}, \textit{MUNIT} \cite{huang2018multimodal}, \textit{DRIT} \cite{lee2018diverse}, \textit{DistanceGAN} and \textit{SelfDistance} \cite{NIPS2017_59b90e10}, and \textit{GCGAN} \cite{Fu_2019_CVPR}). Generative adversarial networks \cite{NIPS2014_5ca3e9b1} have proven to be especially useful in the domain of unsupervised learning i.e.\ when the data used to train the model is not labeled, which is the exact case in this kind of scenario.

The implementation details of this particular approach and what differentiates it from its predecessors will be discussed in more detail in the upcoming chapters.
In \autoref{chap:cut} \textit{Contrastive Unpaired Image-to-Image Translation} \cite{park2020cut} is explained with the \autoref{chap:experiments} showing different experiments and results with proposed loss modifications to the existing models.

\section{Related Work}

\textbf{Generative models.} Ever increasing demand for data has lead to various methods for synthetic data generation. 
The most popular such models that have shown the best results are Generative Adversarial Networks (GANs) \cite{NIPS2014_5ca3e9b1}, Variational Autoencoders (VAEs) \cite{kingma2014autoencoding, vahdat2021nvae} and, more recently, models based on normalizing flows \cite{Kobyzev_2020, sorkhei2020fullglow}. This paper will be focusing on Generative Adversarial models with a specific approach to achieving image-to-image translation.

\textbf{Image-To-Image Translation.} A lot of recently developed models have focused on image-to-image translation, i.e.\ to learn the mapping between an input and an output image in a way that the content of the original input image is preserved while characteristics of output image are applied. This includes \textit{CycleGAN} \cite{CycleGAN2017}, \textit{pix2pix} \cite{isola2018imagetoimage}, \textit{MUNIT} \cite{huang2018multimodal}, \textit{DRIT} \cite{lee2018diverse}, \textit{DistanceGAN} and \textit{SelfDistance} \cite{NIPS2017_59b90e10}, \textit{GCGAN} \cite{Fu_2019_CVPR}, etc. More detailed comparisons and explanations of different implementations available to date can be found in \textit{Image-to-Image Translation: Methods and Applications} \cite{pang2021imagetoimage}.

\textbf{Contrastive Learning.} For quite some time cross-entropy loss was the main loss function, especially in the supervised setting. 
However, recently a new type of loss known as \textit{contrastive loss} has lead to state of the art performance both in supervised \cite{khosla2021supervised} and unsupervised \cite{chen2020a} settings. 
This type of loss allows the model to learn an embedding where associated signals (in our case patches of an image) are brought together while signals that are not associated to each other are pushed further apart in the embedding space\footnote{a relatively low-dimensional space into which high-dimensional vectors are translated}.\\

\section{Contrastive Learning for Unpaired Image-to-Image Translation}
\label{chap:cut}

Image-to-image translation aims to take images from one domain and translate them in such way that they adopt the style or certain characteristics of images from another domain while simultaneously preserving the contents of the original image. This can also be viewed as a disentanglement problem where the goal is to separate the content, which has to be preserved, from the appearance which is supposed to change.

\begin{figure}[htbp]
    \centerline{\includegraphics[scale=0.20]{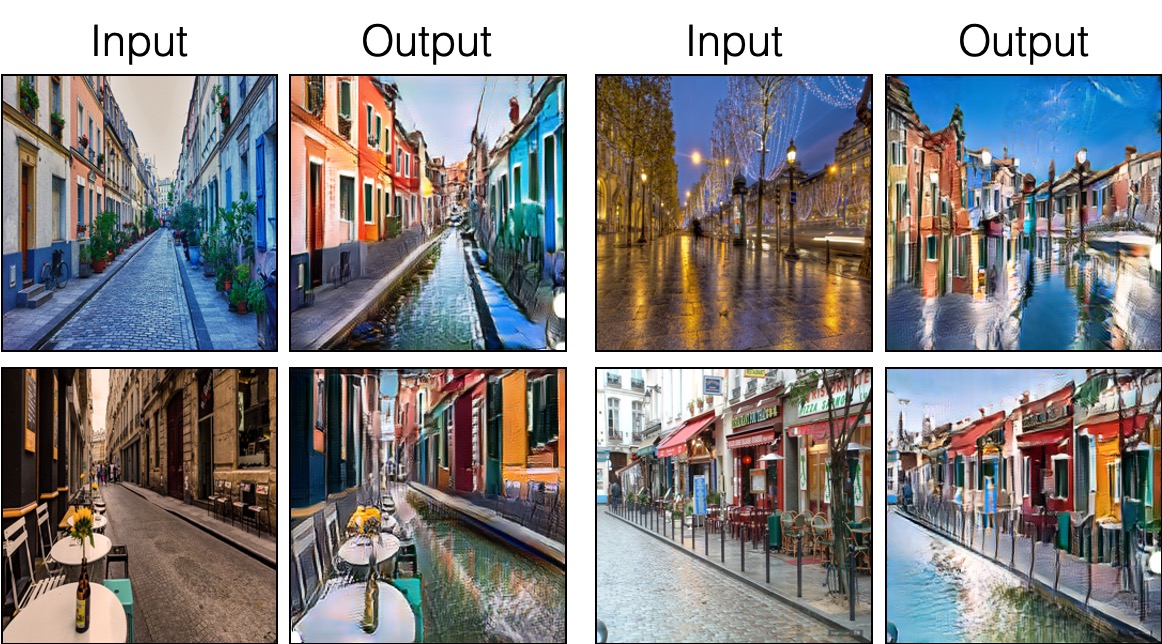}}
    \captionsource{Example of \textbf{Image-to-Image Translation} achieved by the \textbf{CUT model} \cite{park2020cut} - Parisian streets translated to depict the canals of Burano in Venice.}{\cite{park2020cut}}
    \label{fig:paris_burano}
\end{figure}

An example of a successful translation can be seen in the \autoref{fig:paris_burano} where the network was trained using photographs of Parisian streets that represented the input domain and a set of images showing canals of Burano as the output domain.

The upcoming chapters will review a model for performing image-to-image translation, including both its architectural details as well as the approach it takes to overcome the challenges of translation.

\section{Methods}
\label{sec:methods}

Given a dataset of unpaired instances $X$ = \{$x$ $\in$ $\mathcal{X}$\}, $Y$ = \{$y$ $\in$ $\mathcal{Y}$\}, our goal is to translate the images from the input domain $\mathcal{X} \subset \mathbb{R}^{H \times W \times C}$ to appear like images from the output domain $\mathcal{Y} \subset \mathbb{R}^{H \times W \times 3}$ while preserving content of the input domain.

Since this method avoids using inverse auxiliary generators and discriminators, the training procedure is significantly simplified, thus reducing the training times. The generator function \textit{G} consists of two components, an encoder $G_{\textup{enc}}$ followed by a decoder $G_{\textup{dec}}$ which together produce the output image $\hat{\boldsymbol y}=G(\boldsymbol x)=G_{\textup{dec}}(G_{\textup{enc}}(\boldsymbol x))$.

In order to encourage the output to be visually as close as possible to the images contained in the target domain, \textbf{adversarial loss} \cite{NIPS2014_5ca3e9b1} is used:
\begin{multline}
    \label{eqn:adversarial}
    \mathcal{L}_{\textup{GAN}}(G,D,X,Y)=\mathbb{E}_{\boldsymbol{y}\sim Y}\log(D(\boldsymbol{y})) + \\
    \mathbb{E}_{\boldsymbol{x}\sim X}\log(1 - D(G(\boldsymbol{x}))).
\end{multline}

\textbf{Mutual information maximization} is achieved by using noise contrastive estimation \cite{oord2019representation} and learning an embedding such that we closely associate a "query" and its "positive" while disassociating the "query" from other points in the dataset referred to as "negatives".
The query, positive and \textit{N} negatives are mapped to \textit{K}-dimensional vectors $\boldsymbol\upsilon,\boldsymbol\upsilon^+\in\mathbb{R}^K$ and $\boldsymbol\upsilon^-\in\mathbb{R}^{N\times K}$ ($\boldsymbol\upsilon_n^-\in\mathbb{R}^K$ represents the n-th negative), respectively. Thereafter the vectors are normalized onto a unit sphere which prevents the space from collapsing or expanding. Afterwards an (\textit{N}+1)-way classification problem is set up. The distances between the query and other examples (the positive and negatives) are scaled by a temperature $\tau=0.07$ and passed as logits. The loss function is then calculated, which is represented as \textbf{cross-entropy loss} in the original \textit{CUT} model \cite{park2020cut}.

\begin{multline}
    \ell(\boldsymbol{\upsilon,\upsilon^+, \upsilon^-}) = \\ -\log\Bigg[\frac{\exp(\boldsymbol{\upsilon}\cdot\boldsymbol{\upsilon^+}/\tau)}{\exp(\boldsymbol{\upsilon}\cdot\boldsymbol{\upsilon^+}/\tau) + \sum_{n=1}^N\exp(\boldsymbol\upsilon\cdot\boldsymbol{\upsilon_{n}^-}/\tau)}\Bigg].
\end{multline}

Later in \autoref{chap:experiments}, the use of \textbf{focal loss} \cite{lin2018focal} is proposed as an improvement over the classic cross-entropy loss as it shows performance increase in the scenarios where the datasets are unbalanced. 

The ultimate goal is to form association between the input and output data. The query refers to output whereas the positives and negatives are the corresponding and noncorresponding input.
\begin{figure}[t!]
    \centerline{\includegraphics[scale=0.05]{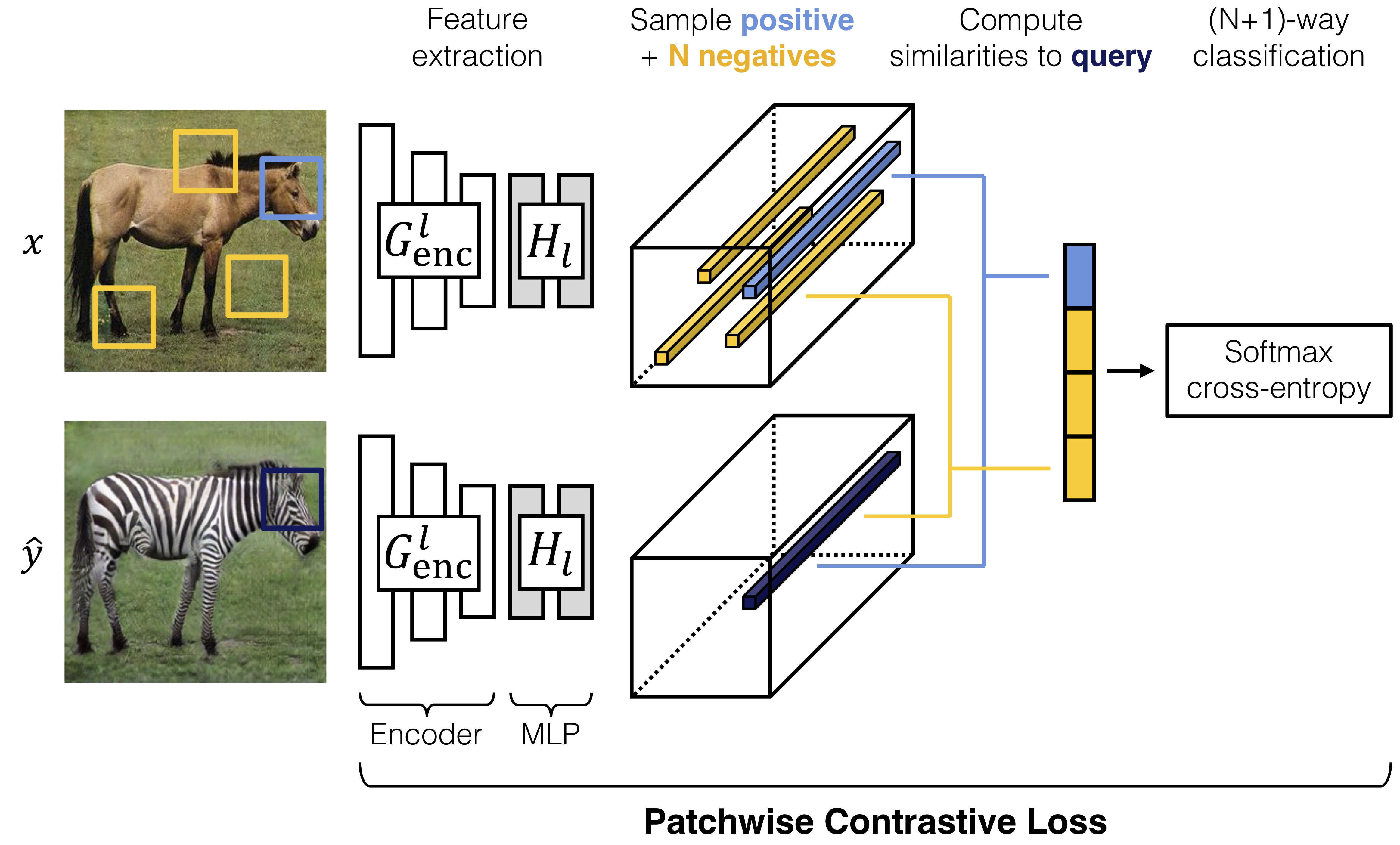}}
    \captionsource{Visual representation of \textbf{Patchwise Contrastive Loss}. Images $x$ and $\hat{y}$ are encoded into feature tensors. A query patch is sampled from the output $\hat{y}$ and compared to a patch (positive) from the input $x$ at the corresponding location. By sampling additional N negatives at different locations in the input we form an (N+1)-way classification problem. The encoder part of the network $G_{\textup{enc}}$ is reused and a two-layer MLP network is added. We use the MLP to produce feature stacks by passing feature maps of layers through it. This procedure projects both the input and the output into a shared embedding space.}{\cite{park2020cut}}
    \label{fig:patchwise_contrastive_loss}
\end{figure}

It is important to notice that not only do we want whole images to share content but also the corresponding patches within the image. 
For instance, in the \textit{Horse}$\to$\textit{Zebra} example we should clearly be able to associate the specific body parts of a zebra to the specific body parts of an input horse more than to the other patches of the input image, e.g.\ a leg of a zebra should be more closely associated to the corresponding leg of a horse than to the other parts of its body.
Moreover, colors of a zebra's body (black and white) are strongly associable to the color of a horse body rather than the background.
Same analogy can be applied in the case of \textit{Satellite}$\to$\textit{Maps} problem covered in \autoref{chap:experiments}. 
Suppose we have a satellite image of an urban environment, after the translation is performed one should clearly be able to associate the streets seen in the \textit{satellite} imagery with the streets visible in the \textit{maps} representation, while disassociating them from other parts e.g.\ bodies of water, green surfaces, etc.
In order to enable achieving this type of correspondence a multi-layer, patch-based objective is set.

Role of the encoder $G_{\textup{enc}}$ is to perform the image translation meaning its feature stack is at our disposal and we can make use of it. 
The feature stack contains layers and spatial locations representing patches of the input image, where deeper layers correspond to larger patches. 
$L$ layers of interest are selected and their feature maps are passed through a two-layer \textit{MLP} (\textit{Multilayer Perceptron}) network $H_l$. 
This procedure produces a stack of features ${\{z_l\}}_L=\{H_l(G_{\textup{enc}}^l(x))\}_L$, where $G_{\textup{enc}}^l$ represents the output of \textit{l}-th chosen layer. 
Layers are then indexed into $l\in\{1,2,...,L\}$ and $s\in\{1,...,S_l\}$ is denoted, where $s$ represents a spatial location and $S_l$ represents the number of spatial locations in each of the layers. 
Corresponding feature is referred to as $z_l^s\in\mathbb{R}^{C_l}$ while $z_l^{S\backslash s}\in\mathbb{R}^{(S_l-1)\times C_l}$ represents the rest of the features, where $C_l$ is the number of channels at each layer. 
The output image $\boldsymbol{\hat{y}}$ is encoded into $\{\hat{z}_l\}_L=\{H_l(G_{\textup{enc}}^l(G_{\textup{dec}}(G_{\textup{enc}}(\boldsymbol x))))\}_L$ in a similar manner.\\
Ultimately, our goal is to match corresponding input-output patches at a specific location. Other patches from within the input image are leveraged as negatives. As previously mentioned, we are aiming to achieve shared content not only between the images as a whole but also the corresponding patches within the image. To achieve this we use \textit{PatchNCE} \cite{park2020cut} loss:

\begin{equation}
    \label{eqn:patchnce}
    \mathcal{L}_{\textup{PatchNCE}}(G,H,X)=\mathbb{E}_{x\sim X}\sum_{l=1}^L\sum_{s=1}^{S_l}\ell(\boldsymbol{\hat{z}_l^s},\boldsymbol{z_l^s},\boldsymbol{z_l^{S\backslash s}}).
\end{equation}


By utilizing all the above mentioned, using \autoref{eqn:adversarial} and \autoref{eqn:patchnce}, we form the following loss function:

\begin{multline}
    \mathcal{L}_{\textup{GAN}}(G,D,X,Y)+\lambda_X\mathcal{L}_{\textup{PatchNCE}}(G,H,X)+ \\
    \lambda_Y\mathcal{L}_{\textup{PatchNCE}}(G,H,Y).
    \label{equation:loss}
\end{multline}

In addition to applying PatchNCE loss on images from domain $\mathcal{X}$ we also apply it on images from domain $\mathcal{Y}$ (referred to as \textit{identity loss}) when using \textit{Contrastive Unpaired Translation (CUT)} model where $\lambda_X=\lambda_Y=1$. This is done in order to prevent the generator from making unnecessary changes - we calculate PatchNCE loss between a translated image and a real image taken from the target domain dataset.
A lighter and faster method called \textit{FastCUT} is also provided in which case $\lambda_X=10$ in order to compensate for absence of the regularizer, i.e.\ for $\lambda_Y=0$. In the experimental phase we show that although the model is faster and lighter to train, the absence of the regularizer can result in significant oscillations in model's performance during the training and potentially cause it to achieve subpar final results. \textit{FastCUT} is designed for usage when time to train is restricted or GPU memory limitations are present as it can achieve satisfactory results similar to \textit{CycleGAN} \cite{CycleGAN2017}, while requiring significantly less memory and time to train.

\section{Datasets}

\subsection{Maps Dataset}

\textbf{Sat2map dataset} contains test, train and validation sets each containing 1098, 1096, and 1098 images, respectively. The dimensions of images are $600\times 600$ with each image containing its representation in both domains, i.e.\ in input domain $\mathcal{A}$ and output domain $\mathcal{B}$. Although the images in the dataset are inherently paired, we preserve the nature of the unpaired approach as the images are not taken in corresponding pairs, i.e.\ the source and target datasets are unaligned, during the training process. Domain $\mathcal{A}$ consists of various satellite images, predominantly those containing dense urban areas, while domain $\mathcal{B}$ contains corresponding representations in the style of street maps.

\subsection{Horses to Zebras Dataset}

\textbf{Horse2zebra} dataset consists of a test set containing 120 horse and 140 zebra images, and a train set containing 1067 horse images and 1334 zebra images, respectively. Both datasets are unpaired. Resolution of each image is $256 \times 256$.

\section{Experiments}
\label{chap:experiments}

\subsection{Evaluation Procedure}

In order to evaluate the quality of generated images we use \textbf{Fréchet Inception Distance (FID)} \cite{Seitzer2020FID, heusel2018gans} metric:

\begin{equation}
    \textup{FID} = \|\mu_X - \mu_Y\|^2 + \textup{tr}\Big(\Sigma_X + \Sigma_Y - 2\sqrt{(\Sigma_X \Sigma_Y)}\Big).
\end{equation}

\noindent In short, what this method does is empirically estimates the distribution of real and generated images in a deep network space and computes the divergence between them.
Ideally, we want to have the lowest possible FID score, which indicates that the generated images are convincing.

\subsection{Training Details}

This section covers the performance of \textit{CUT} and \textit{FastCUT} models with different losses against \textit{CycleGAN} on \textit{sat2map} dataset.\\
The initial \textit{CUT} model includes ResNet-based generator \cite{johnson2016perceptual} with 9 residual blocks, PatchGAN discriminator \cite{isola2018imagetoimage}, Least SquareGAN loss \cite{mao2017squares}, batch size of 1 and Adam optimizer \cite{kingma2017adam} with learning rate of 0.0002 - this is in-line with the original settings provided in \textit{Contrastive Learning for Unpaired Image-to-Image Translation} \cite{park2020cut} paper. 
The reason we choose such parameters is to be as close as possible to \textit{CycleGAN} \cite{CycleGAN2017}.
The only difference is is that the $\ell_1$ cycle-consistency loss\footnote{forward cycle-consistency loss: $x \rightarrow G(x) \rightarrow F(G(x)) \approx x$ \newline \indent \space \space \space backward cycle-consistency loss: $y \rightarrow F(y) \rightarrow G(F(y)) \approx y$} (\cite{CycleGAN2017}) is replaced with contrastive loss \cite{park2020cut}.\\
As previously mentioned, in \autoref{sec:methods} for \textit{CUT} model we use $\lambda_X = \lambda_Y=1$ whereas for \textit{FastCUT} $\lambda_X=10$ and $\lambda_Y$ = 0 is used in the loss function (\autoref{equation:loss}).\\
Each \textit{CUT} experiment is trained up to 400 epochs, where during first 200 epochs the learning rate is kept constant and throughout the last 200 epochs it gradually decays to 0. Moreover, \textit{FastCUT} model is trained up to 200 epochs with first 150 epochs keeping a constant learning rate of 0.0002 and last 50 decaying it at a constant rate. Additionally, flip-equivariance augmentation is applied when training \textit{FastCUT} as described in the original paper \cite{park2020cut}. For calculating PatchNCE loss we extract features from 5 different layers of the $G_{\textup{enc}}$. Namely, RGB pixels, the first and second downsampling convolution as well as the first and the fifth residual block. These layers correspond to receptive fields of sizes $1 \times 1$, $9 \times 9$, $15 \times 15$, $35 \times 35$ and $99 \times 99$. For each layer's features, a 2-layer MLP is applied onto 256 randomly sampled locations in order to acquire 256-dim final features.

\subsection{Results}

\begin{figure}[htbp]
    \begin{subfigure}{\textwidth}
        \includegraphics[width=.49\linewidth]{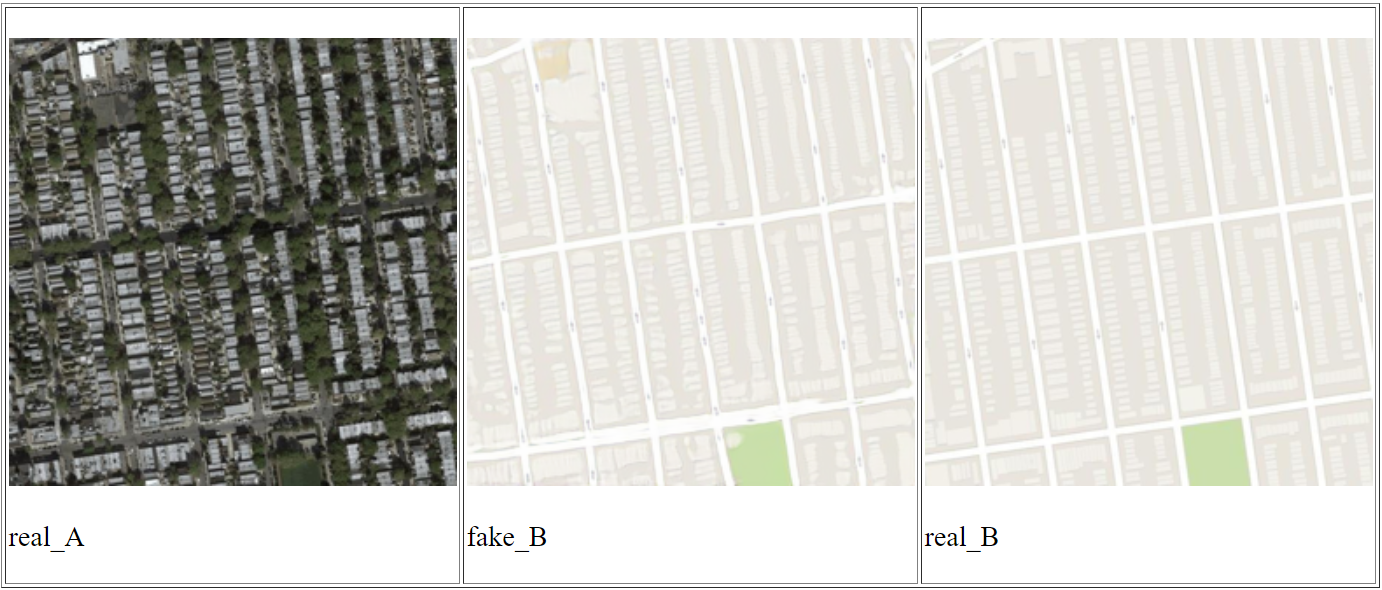}
    \end{subfigure}
    \begin{subfigure}{\textwidth}
        \includegraphics[width=.49\linewidth]{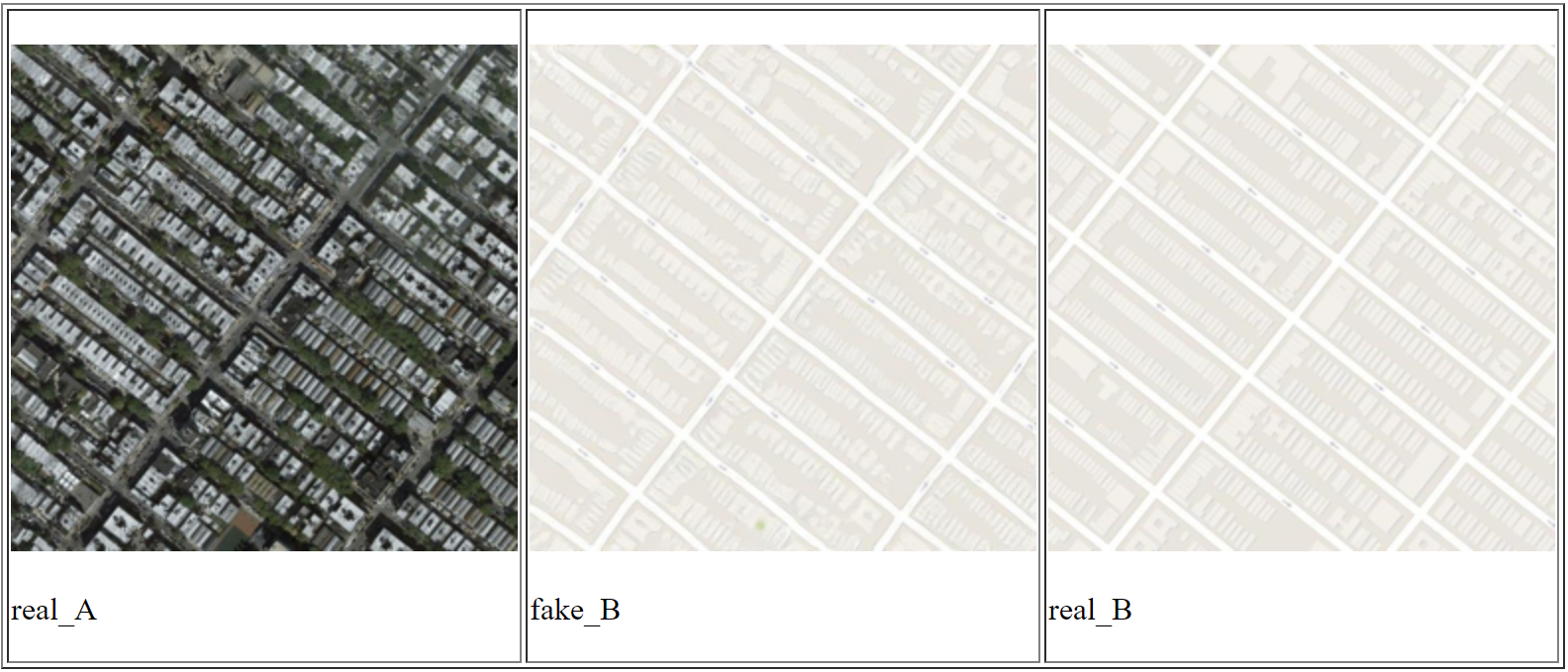}
    \end{subfigure}
    \centering
    \caption{\textbf{Example of a successful $\textit{satellite}\rightarrow\textit{map}$ translation} achieved by \textit{CUT - CE} model (Target domain (domain $\mathcal{B}$) is represented by the image labeled \textit{real\_B} while the image labeled \textit{real\_A} represents the source domain (domain $\mathcal{A}$). The result of translation is shown in the middle image labeled \textit{fake\_B}.)}
    \label{fig:good_example}
\end{figure}

\begin{figure}[htbp]
    \begin{subfigure}{\textwidth}
        \includegraphics[width=.49\linewidth]{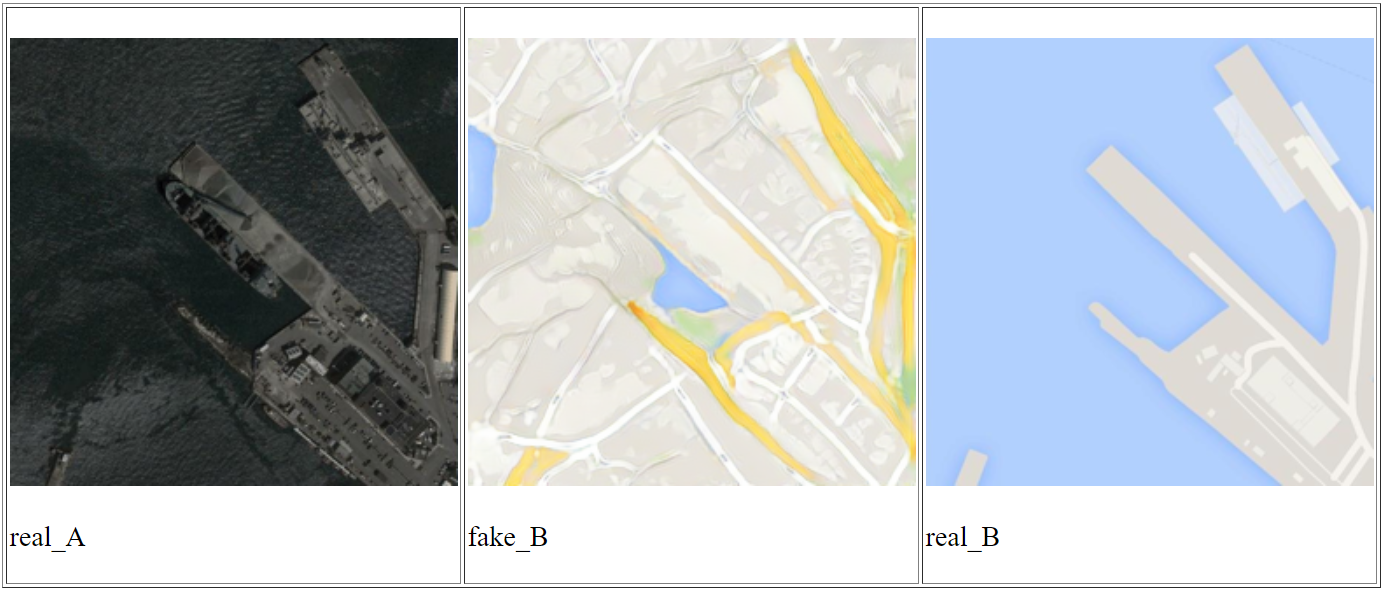}
    \end{subfigure}
    \begin{subfigure}{\textwidth}
        \includegraphics[width=.49\linewidth]{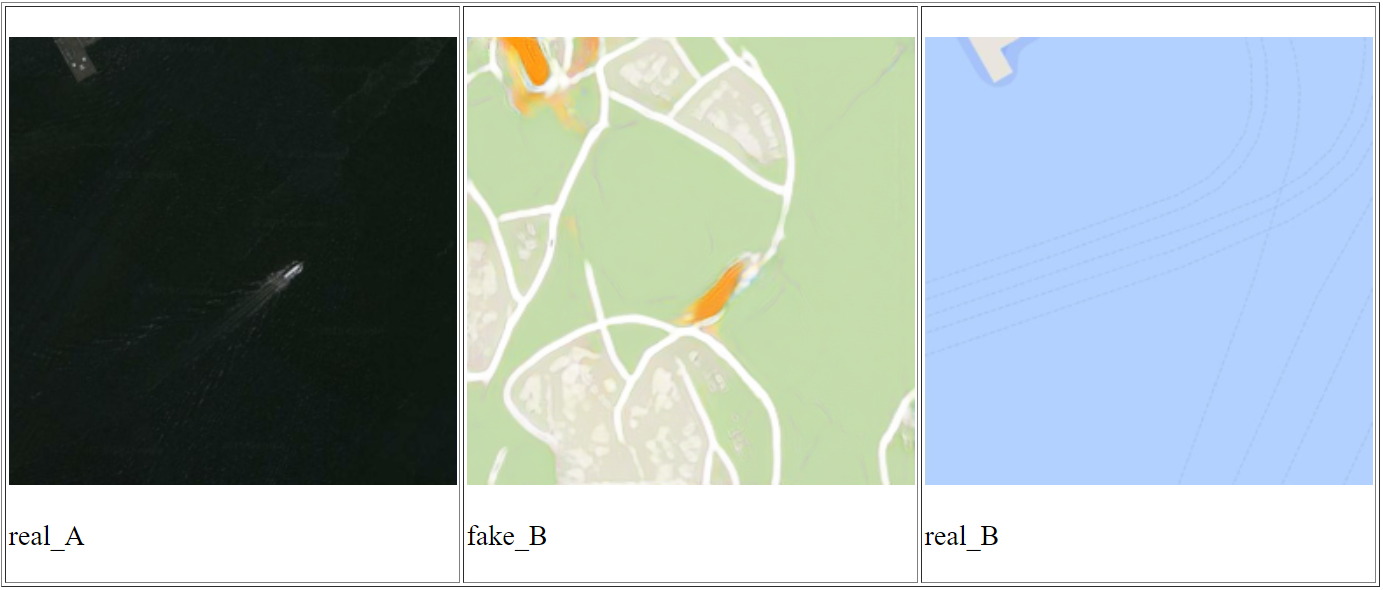}
    \end{subfigure}
    \centering
    \caption{\textbf{Example of an edge case $\textit{satellite}\rightarrow\textit{map}$ translation} achieved by \textit{CUT - CE} model where the output is severely artifacted. (Target domain (domain $\mathcal{B}$) is represented by the image labeled \textit{real\_B} while the image labeled \textit{real\_A} represents the source domain (domain $\mathcal{A}$). The result of translation is shown in the middle image labeled \textit{fake\_B}.)}
    \label{fig:bad_example}
\end{figure}

For the first experiment we use \textit{sat2map} dataset and train the \textit{CUT} model without any modifications using methodology explained above, we will refer to it as \textit{CUT - CE}.
While this model already surpasses the performance of \textit{CycleGAN} \cite{CycleGAN2017} by $\sim$5\% in terms of FID score (\autoref{tab:fid_scores}) and is quite good at translating satellite imagery containing exclusively urban environments (\autoref{fig:good_example}), severe artifacts and completely incorrect translations can be seen in some edge cases, e.g.\ in images that contain bodies of water or green surfaces (\autoref{fig:bad_example}).\\
This is most likely caused by the imbalance of the dataset as it contains a plethora of images containing urban environments while it lacks satellite imagery of green surfaces, bodies of water and other similar edge cases that cause problems in translation.

\begin{figure}[htbp]
    \centering
    \includegraphics[scale=0.15]{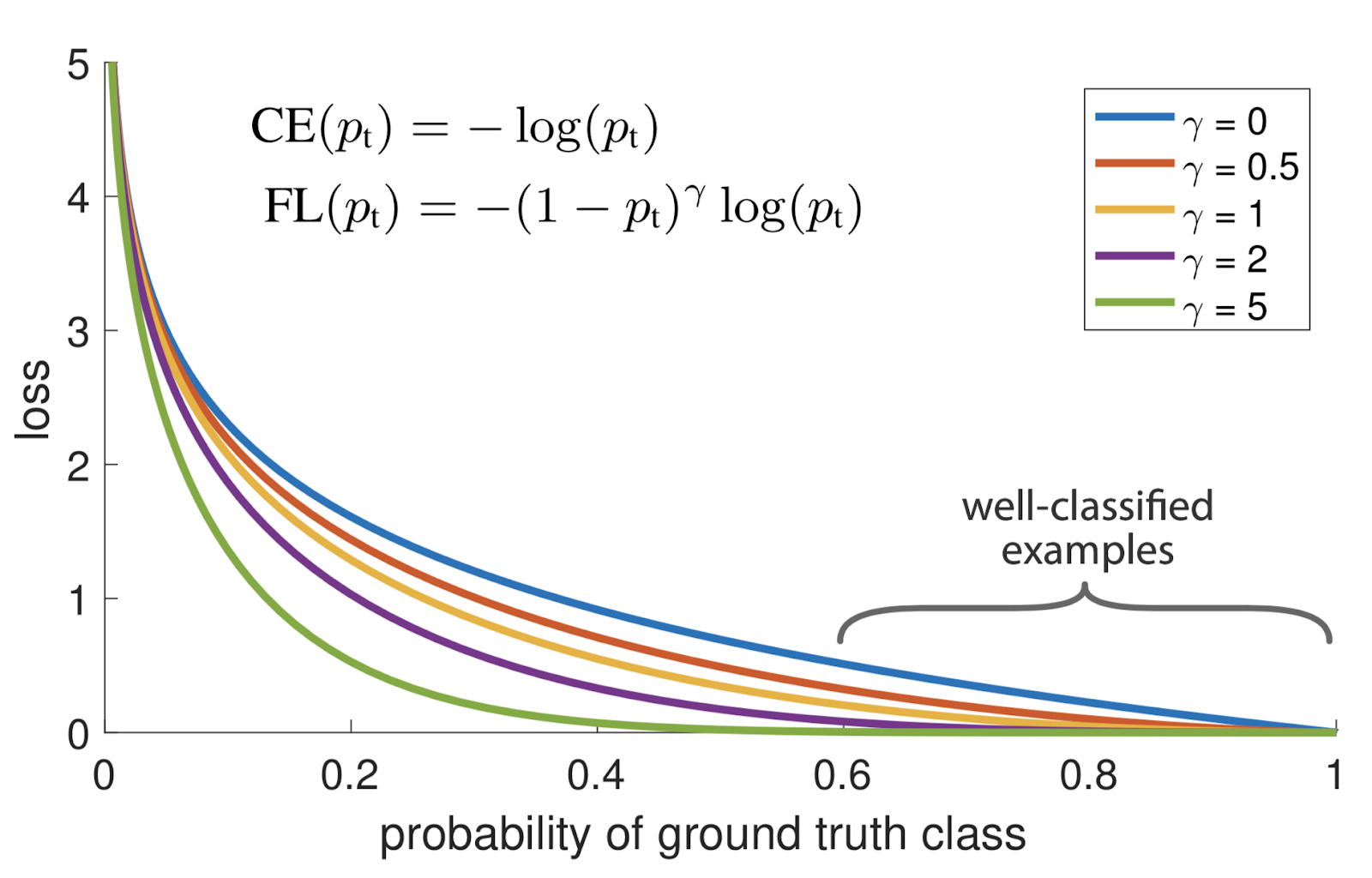}
    \captionsource{\textbf{Visual representation of focal loss} \cite{lin2018focal} depending on different values of $\gamma$. Focal loss with $\gamma =0$ is a cross-entropy loss.}{\cite{lin2018focal}}
    \label{fig:focal_loss}
\end{figure}

\noindent In order to alleviate this issue we introduce \textbf{focal loss} \cite{lin2018focal} (\autoref{fig:focal_loss}):

\begin{equation}
    \textup{FL}(p_t) = -\alpha(1-p_t)^\gamma \log(p_t)
\end{equation}

\noindent This is done by replacing previously used cross-entropy loss within the \textit{PatchNCE} loss while keeping other specified training parameters.\\
Focal loss allows us to penalize difficult, misclassified examples while significantly reducing the loss for well classified examples. 
Intuitively, this will allow our model to focus on edge case scenarios and adjust accordingly to increase its overall performance while making minimal changes in the scenarios where its performance is already good.\\

\noindent We proceed to train the model using focal loss with $\gamma = 2$ and $\alpha = 0.25$ and achieve additional $\sim$3\% boost in FID score compared to regular cross-entropy model, bringing the total performance increase compared to \textit{CycleGAN} up to $\sim$8\%.
An additional model is trained with $\gamma = 3$ and $\alpha = 0.5$, and, while it still outperforms \textit{CycleGAN} and \textit{CUT - CE}, it fails to outperform the initial configuration using $\gamma = 2$ and $\alpha = 0.25$.\\
It is clearly visible in \autoref{tab:fid_scores} that all the tested \textit{CUT} models use significantly less memory (almost $3\times$ less) than \textit{CycleGAN}, train faster and simultaneously manage to outperform \textit{CycleGAN's} FID scores.\\
Furthermore, same experiment is performed using the \textit{FastCUT} model. First it's trained using Cross-Entropy loss (denoted as \textit{FastCUT - CE}), and afterwards using focal loss. 
As shown in \autoref{tab:fid_scores} focal loss version significantly outperforms cross-entropy version and even comes close to \textit{CycleGAN} in terms of FID, all while consuming over $4\times$ less memory and taking $\sim$40\% less time per iteration which brings down the total training time from $\sim$37 hours to just $\sim$11 hours.\\
Next we proceed to test the model on the \textit{horse2zebra} dataset as well. Again, the newly proposed method that uses focal loss yields significant improvements (\autoref{tab:fid_scores2}).

\begin{figure}
    \centering
    \includegraphics[scale=0.37]{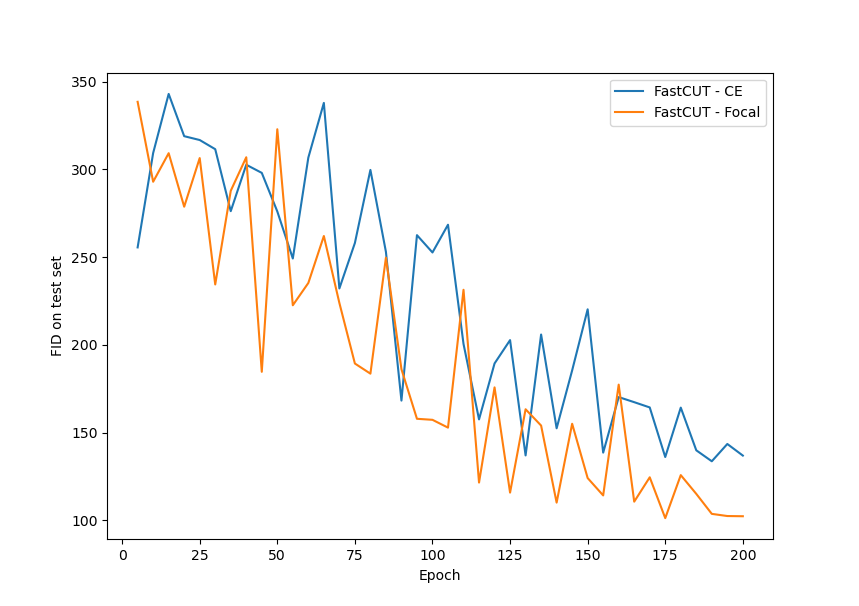}
    \caption{FID scores of \textit{FastCUT} variants on test set each 5 epochs throughout 200 epochs. While the performance of both models oscillates due to the missing identity loss, version using focal loss converges faster reaching a lower final FID score.}
    \label{fig:my_label}
\end{figure}

\begin{table}[htbp]
    \centering
    \caption{\textbf{Comparison of different methods used on sat2map dataset}. FID is calculated using fid-pytorch library \cite{Seitzer2020FID}. It is calculated between domain $\mathcal{B}$ test dataset in \textit{sat2map} and 500 generated images using a trained model. Testing was done using an NVIDIA Tesla P100 using Google Colab platform.}
    \begin{tabular}{@{}lccc@{}}
    \toprule
    \textbf{Method}      & \textbf{FID $\downarrow$}   & \textbf{sec/iter $\downarrow$} & \textbf{Mem(GiB) $\downarrow$} \\ \midrule
    CycleGAN    & \multicolumn{1}{c}{99.28} &   0.33   & 9.48     \\ \hdashline
    CUT - CE        & 94.48                     &   0.3   &   3.39  \\
    CUT - Focal ($\gamma = 2, \alpha = 0.25$) & \textbf{91.44}                     &    0.3  &   3.39 \\
    CUT - Focal ($\gamma = 3, \alpha = 0.5$) & 93.24                     &  0.3    &  3.39  \\ \hdashline
    FastCUT - CE & 136.98                   & 0.19     & 2.34   \\
    FastCUT - Focal ($\gamma = 2, \alpha = 0.25)$ & \textbf{102.41}                & 0.19     & 2.34   \\ \bottomrule
    \end{tabular}
    \label{tab:fid_scores}
\end{table}

\begin{table}[htbp]
    \centering
    \caption{\textbf{Comparison of different methods used on horse2zebra dataset.}}
    \begin{tabular}{@{}lccc@{}}
    \toprule
    \textbf{Method}      & \textbf{FID $\downarrow$}   & \textbf{sec/iter $\downarrow$} & \textbf{Mem(GiB) $\downarrow$} \\ \midrule
    CycleGAN    & \multicolumn{1}{c}{77.2} &   0.40   & 4.81     \\ 
    \textbf{DCLGAN (current SOTA)}    & \multicolumn{1}{c}{43.2} &   0.41   & n/a     \\ \hdashline
    CUT - CE        & 45.5                     &   0.24   &   3.33  \\
    \textbf{CUT - Focal ($\gamma = 2, \alpha = 0.25$)} & \textbf{41.8}                     &    0.24  &   3.33 \\
    \bottomrule
    \end{tabular}
    \label{tab:fid_scores2}
\end{table}

\newpage
\section{Conclusion}

The main focus of this paper was to explore potential improvements of pre-existing Contrastive Unpaired Translation method.
This approach focuses on maximizing mutual information between corresponding input and output patches in order to form the final output image which contains the content of input domain and the characteristics of output domain while using focal loss for patch classification.

By conducting a series of experiments on \textit{sat2map} and \textit{horse2zebra} datasets using different variations of the \textit{CUT} model we compare their performances based on FID, training speeds and memory usage. Based on the  results, an upgraded version of \textit{CUT} model which uses focal loss instead of cross-entropy loss within \textit{PatchNCE} loss is proposed as it has the ability to alleviate the imbalance issue of \textit{sat2map} and \textit{horse2zebra} datasets and improve the convergence of \textit{FastCUT} model.

\section{Acknowledgments}

The experiments presented in this paper were conducted as a part of my bachelor's thesis at the Faculty of Electrical Engineering and Computing, University of Zagreb under supervision of Prof. Siniša Šegvić.

\bibliographystyle{IEEEtran}
\bibliography{IEEEabrv,bibliography}

\begin{thebibliography}{10}
\providecommand{\url}[1]{#1}
\csname url@samestyle\endcsname
\providecommand{\newblock}{\relax}
\providecommand{\bibinfo}[2]{#2}
\providecommand{\BIBentrySTDinterwordspacing}{\spaceskip=0pt\relax}
\providecommand{\BIBentryALTinterwordstretchfactor}{4}
\providecommand{\BIBentryALTinterwordspacing}{\spaceskip=\fontdimen2\font plus
\BIBentryALTinterwordstretchfactor\fontdimen3\font minus
  \fontdimen4\font\relax}
\providecommand{\BIBforeignlanguage}[2]{{%
\expandafter\ifx\csname l@#1\endcsname\relax
\typeout{** WARNING: IEEEtran.bst: No hyphenation pattern has been}%
\typeout{** loaded for the language `#1'. Using the pattern for}%
\typeout{** the default language instead.}%
\else
\language=\csname l@#1\endcsname
\fi
#2}}
\providecommand{\BIBdecl}{\relax}
\BIBdecl

\bibitem{park2020cut}
\BIBentryALTinterwordspacing
T.~Park, A.~A. Efros, R.~Zhang, and J.~Zhu, ``Contrastive learning for unpaired
  image-to-image translation,'' \emph{CoRR}, vol. abs/2007.15651, 2020.
  [Online]. Available: \url{https://arxiv.org/abs/2007.15651}
\BIBentrySTDinterwordspacing

\bibitem{CycleGAN2017}
\BIBentryALTinterwordspacing
J.~Zhu, T.~Park, P.~Isola, and A.~A. Efros, ``Unpaired image-to-image
  translation using cycle-consistent adversarial networks,'' \emph{CoRR}, vol.
  abs/1703.10593, 2017. [Online]. Available:
  \url{http://arxiv.org/abs/1703.10593}
\BIBentrySTDinterwordspacing

\bibitem{NIPS2014_5ca3e9b1}
\BIBentryALTinterwordspacing
I.~Goodfellow, J.~Pouget-Abadie, M.~Mirza, B.~Xu, D.~Warde-Farley, S.~Ozair,
  A.~Courville, and Y.~Bengio, ``Generative adversarial nets,'' in
  \emph{Advances in Neural Information Processing Systems}, Z.~Ghahramani,
  M.~Welling, C.~Cortes, N.~Lawrence, and K.~Q. Weinberger, Eds.,
  vol.~27.\hskip 1em plus 0.5em minus 0.4em\relax Curran Associates, Inc.,
  2014. [Online]. Available:
  \url{https://proceedings.neurips.cc/paper/2014/file/5ca3e9b122f61f8f06494c97b1afccf3-Paper.pdf}
\BIBentrySTDinterwordspacing

\bibitem{huang2018multimodal}
\BIBentryALTinterwordspacing
X.~{Huang}, M.-Y. {Liu}, S.~J. {Belongie}, and J.~{Kautz}, ``Multimodal
  unsupervised image-to-image translation,'' in \emph{Proceedings of the
  European Conference on Computer Vision (ECCV)}, 2018, pp. 179--196. [Online].
  Available: \url{https://arxiv.org/pdf/1804.04732.pdf}
\BIBentrySTDinterwordspacing

\bibitem{lee2018diverse}
\BIBentryALTinterwordspacing
H.~Lee, H.~Tseng, J.~Huang, M.~K. Singh, and M.~Yang, ``Diverse image-to-image
  translation via disentangled representations,'' \emph{CoRR}, vol.
  abs/1808.00948, 2018. [Online]. Available:
  \url{http://arxiv.org/abs/1808.00948}
\BIBentrySTDinterwordspacing

\bibitem{NIPS2017_59b90e10}
\BIBentryALTinterwordspacing
S.~Benaim and L.~Wolf, ``One-sided unsupervised domain mapping,'' in
  \emph{Advances in Neural Information Processing Systems}, I.~Guyon, U.~V.
  Luxburg, S.~Bengio, H.~Wallach, R.~Fergus, S.~Vishwanathan, and R.~Garnett,
  Eds., vol.~30.\hskip 1em plus 0.5em minus 0.4em\relax Curran Associates,
  Inc., 2017. [Online]. Available:
  \url{https://proceedings.neurips.cc/paper/2017/file/59b90e1005a220e2ebc542eb9d950b1e-Paper.pdf}
\BIBentrySTDinterwordspacing

\bibitem{Fu_2019_CVPR}
H.~Fu, M.~Gong, C.~Wang, K.~Batmanghelich, K.~Zhang, and D.~Tao,
  ``Geometry-consistent generative adversarial networks for one-sided
  unsupervised domain mapping,'' in \emph{Proceedings of the IEEE/CVF
  Conference on Computer Vision and Pattern Recognition (CVPR)}, June 2019.

\bibitem{kingma2014autoencoding}
\BIBentryALTinterwordspacing
D.~P. Kingma and M.~Welling, ``Auto-encoding variational bayes,'' 2014.
  [Online]. Available: \url{https://arxiv.org/pdf/1312.6114.pdf}
\BIBentrySTDinterwordspacing

\bibitem{vahdat2021nvae}
\BIBentryALTinterwordspacing
A.~Vahdat and J.~Kautz, ``Nvae: A deep hierarchical variational autoencoder,''
  2021. [Online]. Available: \url{https://arxiv.org/pdf/2007.03898.pdf}
\BIBentrySTDinterwordspacing

\bibitem{Kobyzev_2020}
\BIBentryALTinterwordspacing
I.~Kobyzev, S.~Prince, and M.~Brubaker, ``Normalizing flows: An introduction
  and review of current methods,'' \emph{IEEE Transactions on Pattern Analysis
  and Machine Intelligence}, p. 1–1, 2020. [Online]. Available:
  \url{http://dx.doi.org/10.1109/TPAMI.2020.2992934}
\BIBentrySTDinterwordspacing

\bibitem{sorkhei2020fullglow}
\BIBentryALTinterwordspacing
M.~Sorkhei, G.~E. Henter, and H.~Kjellstr{\"{o}}m, ``Full-glow: Fully
  conditional glow for more realistic image generation,'' \emph{CoRR}, vol.
  abs/2012.05846, 2020. [Online]. Available:
  \url{https://arxiv.org/abs/2012.05846}
\BIBentrySTDinterwordspacing

\bibitem{isola2018imagetoimage}
\BIBentryALTinterwordspacing
P.~Isola, J.-Y. Zhu, T.~Zhou, and A.~Efros, ``Image-to-image translation with
  conditional adversarial networks,'' 07 2017, pp. 5967--5976. [Online].
  Available: \url{https://arxiv.org/pdf/1611.07004.pdf}
\BIBentrySTDinterwordspacing

\bibitem{pang2021imagetoimage}
\BIBentryALTinterwordspacing
Y.~Pang, J.~Lin, T.~Qin, and Z.~Chen, ``Image-to-image translation: Methods and
  applications,'' \emph{CoRR}, vol. abs/2101.08629, 2021. [Online]. Available:
  \url{https://arxiv.org/abs/2101.08629}
\BIBentrySTDinterwordspacing

\bibitem{khosla2021supervised}
\BIBentryALTinterwordspacing
P.~Khosla, P.~Teterwak, C.~Wang, A.~Sarna, Y.~Tian, P.~Isola, A.~Maschinot,
  C.~Liu, and D.~Krishnan, ``Supervised contrastive learning,'' \emph{CoRR},
  vol. abs/2004.11362, 2020. [Online]. Available:
  \url{https://arxiv.org/abs/2004.11362}
\BIBentrySTDinterwordspacing

\bibitem{chen2020a}
\BIBentryALTinterwordspacing
T.~{Chen}, S.~{Kornblith}, M.~{Norouzi}, and G.~{Hinton}, ``A simple framework
  for contrastive learning of visual representations,'' \emph{arXiv preprint
  arXiv:2002.05709}, 2020. [Online]. Available:
  \url{https://arxiv.org/pdf/2002.05709.pdf}
\BIBentrySTDinterwordspacing

\bibitem{oord2019representation}
\BIBentryALTinterwordspacing
A.~van~den Oord, Y.~Li, and O.~Vinyals, ``Representation learning with
  contrastive predictive coding,'' \emph{CoRR}, vol. abs/1807.03748, 2018.
  [Online]. Available: \url{http://arxiv.org/abs/1807.03748}
\BIBentrySTDinterwordspacing

\bibitem{lin2018focal}
\BIBentryALTinterwordspacing
T.~Lin, P.~Goyal, R.~B. Girshick, K.~He, and P.~Doll{\'{a}}r, ``Focal loss for
  dense object detection,'' \emph{CoRR}, vol. abs/1708.02002, 2017. [Online].
  Available: \url{http://arxiv.org/abs/1708.02002}
\BIBentrySTDinterwordspacing

\bibitem{Seitzer2020FID}
M.~Seitzer, ``{pytorch-fid: FID Score for PyTorch},''
  \url{https://github.com/mseitzer/pytorch-fid}, August 2020, version 0.1.1.

\bibitem{heusel2018gans}
\BIBentryALTinterwordspacing
M.~Heusel, H.~Ramsauer, T.~Unterthiner, B.~Nessler, and S.~Hochreiter, ``Gans
  trained by a two time-scale update rule converge to a local nash
  equilibrium,'' in \emph{Advances in Neural Information Processing Systems},
  I.~Guyon, U.~V. Luxburg, S.~Bengio, H.~Wallach, R.~Fergus, S.~Vishwanathan,
  and R.~Garnett, Eds., vol.~30.\hskip 1em plus 0.5em minus 0.4em\relax Curran
  Associates, Inc., 2017. [Online]. Available:
  \url{https://proceedings.neurips.cc/paper/2017/file/8a1d694707eb0fefe65871369074926d-Paper.pdf}
\BIBentrySTDinterwordspacing

\bibitem{johnson2016perceptual}
\BIBentryALTinterwordspacing
J.~Johnson, A.~Alahi, and F.~Li, ``Perceptual losses for real-time style
  transfer and super-resolution,'' \emph{CoRR}, vol. abs/1603.08155, 2016.
  [Online]. Available: \url{http://arxiv.org/abs/1603.08155}
\BIBentrySTDinterwordspacing

\bibitem{mao2017squares}
\BIBentryALTinterwordspacing
X.~Mao, Q.~Li, H.~Xie, R.~Y.~K. Lau, and Z.~Wang, ``Multi-class generative
  adversarial networks with the {L2} loss function,'' \emph{CoRR}, vol.
  abs/1611.04076, 2016. [Online]. Available:
  \url{http://arxiv.org/abs/1611.04076}
\BIBentrySTDinterwordspacing

\bibitem{kingma2017adam}
\BIBentryALTinterwordspacing
D.~P. Kingma and J.~Ba, ``Adam: {A} method for stochastic optimization,'' in
  \emph{3rd International Conference on Learning Representations, {ICLR} 2015,
  San Diego, CA, USA, May 7-9, 2015, Conference Track Proceedings}, Y.~Bengio
  and Y.~LeCun, Eds., 2015. [Online]. Available:
  \url{http://arxiv.org/abs/1412.6980}
\BIBentrySTDinterwordspacing

\end{thebibliography}

\end{document}